\documentclass[11pt]{article}

\usepackage[final]{acl}
\usepackage{amsmath,amssymb}
\usepackage{algorithm}
\usepackage{algpseudocode}
\usepackage[most]{tcolorbox}
\usepackage{fvextra} % better verbatim
\usepackage{booktabs}
\usepackage{adjustbox}
\usepackage{graphicx}
\usepackage{tcolorbox}

\usepackage{times}
\usepackage{latexsym}

\usepackage[T1]{fontenc}
\usepackage[utf8]{inputenc}

\usepackage{microtype}

\usepackage{inconsolata}

\usepackage{graphicx}

\title{Early-Token Confidence Predicts Reasoning Quality in Multi-Agent LLM Debate}

\author{Ali Keramati, Justin Cheok , Jacob Horne and Mark Warschauer \\
University of California, Irvine\\
\texttt{\{a.kera,jcheok,jhorne1,markw\}@uci.edu} 
}

\begin{document}
\maketitle
\begin{abstract}
Evaluating reasoning quality in multi-agent LLM systems is challenging, especially for open-ended tasks without reference answers. We investigate whether intrinsic confidence signals, token-level log-probabilities from decoding, can predict reasoning quality as assessed by LLM-as-judge evaluation. Using a debate-based essay scoring framework, we compare confidence proxies against rubric-based judge scores across two ASAP essay sets. We find that early-token confidence, particularly within the first few generated tokens, is consistently the strongest predictor of reasoning quality, outperforming full-sequence statistics. Analysis of log-probability trajectories shows that the opening phase of generation is the most heterogeneous and therefore most informative. We also observe a systematic asymmetry between agent roles, with stronger alignment between confidence and quality for supportive reasoning than for adversarial critique. These results suggest that early decoding dynamics provide a lightweight and effective signal for estimating reasoning reliability in multi-agent LLM systems.
\end{abstract}

\section{Introduction}

Recent advances in large language models (LLMs) have enabled the development of \emph{multi-agent systems}, in which multiple specialized agents collaborate to solve complex tasks \cite{wu2023autogenenablingnextgenllm}. By decomposing problems into role-specific subtasks, such systems have been shown to improve performance, robustness, and consistency across a range of applications, including reasoning, planning, and automated decision-making \cite{parmar2025plangenmultiagentframeworkgenerating, han2026llmmultiagentsystemschallenges}. Among interaction paradigms, \emph{debate} has emerged as a particularly effective mechanism: by eliciting both supporting and opposing arguments, it encourages exploration of diverse reasoning paths and exposes errors that may remain hidden in a single-agent trajectory \cite{du2023improvingfactualityreasoninglanguage}.

Rubric-based scoring provides a concrete and high-impact setting in which these benefits are especially relevant. In this setting, a system assigns scores according to a predefined rubric that specifies evaluation criteria and score ranges \cite{10874775}. A canonical example is \emph{automated essay scoring (AES)}, where models aim to approximate human judgments of student writing quality \cite{Dikli_2006}. Public benchmarks such as the ASAP \footnote{https://www.kaggle.com/c/asap-aes/data} dataset include prompts that provide trait-level rubric scores (rather than a single holistic score), enabling trait-specific feedback and analysis \cite{CROSSLEY2025100954}. At the same time, recent work has explored using LLMs directly for essay scoring, highlighting both the promise of scalable rubric-based evaluation and the need to better understand the reliability of LLM-driven scoring behavior \cite{PACK2024100234}.

Multi-agent debate is a natural fit for rubric scoring because it produces inspectable intermediate reasoning artifacts \cite{keramati_2025_17196206}. In these systems, agents adopt complementary roles, generating diverse perspectives on the same input. This structured disagreement can help the system consider alternative interpretations of the rubric and mitigate single-path scoring bias by forcing explicit engagement with counterevidence \cite{du2023improvingfactualityreasoninglanguage}. However, debate also increases system complexity: multiple agents, multiple messages, and multiple opportunities for subtle procedural failures \cite{wynn2025talkisntcheapunderstanding}. As multi-agent pipelines become more elaborate, it becomes essential to add an evaluation layer that measures not only whether a final score matches a reference, but also whether agents' reasoning is high quality and reliable \cite{chen2025multiagentasjudgealigningllmagentbasedautomated}.

A growing body of work addresses this need through \emph{LLM-as-judge} evaluation, where a separate language model is used to score generated outputs along predefined criteria. This paradigm has become a scalable alternative to human evaluation, particularly for open-ended tasks where reference answers are unavailable \cite{zheng2023judging}. However, LLM-as-judge provides only an \emph{external} signal of quality, and an important open question remains: \textit{to what extent do these judgments reflect the true reliability of the underlying reasoning process?} In particular, can we identify \emph{intrinsic signals} within the generating model that correlate with externally judged reasoning quality?

To connect judge-based reasoning evaluation with model-intrinsic signals, we turn to \emph{confidence estimation} and \emph{uncertainty quantification} \cite{kang2025scalable}. Neural probabilities are not automatically calibrated, and language model confidence can be misaligned with correctness. Nevertheless, recent research shows that language models can provide meaningful self-evaluations under appropriate formats and that uncertainty estimation for LLM generation is an active area of study \cite{mavi2025selfevaluatingllmsmultisteptasks}. In this work, we operationalize model confidence using token-level log-probabilities produced during decoding. Intuitively, if an agent follows a more coherent and evidentially grounded reasoning path, the model should assign higher probability mass to the tokens it generates along that path, yielding more confident logprob trajectories.

\section{Related Work}

\paragraph{LLM-as-Judge Evaluation.}
Recent work has established \emph{LLM-as-judge} as a practical paradigm for evaluating open-ended generation in settings where reference answers are weak or unavailable. Prior studies show that strong language models can correlate well with human judgments on instruction-following and related tasks, making them a scalable alternative to manual evaluation \cite{chiang2023can, li2023alpacaeval, zheng2023judging, fu2023gptscore, liu2023geval}. Beyond coarse pairwise or scalar judgments, subsequent work emphasizes the need for more structured and interpretable evaluation. For example, FLASK introduces fine-grained, rubric-based assessment and demonstrates improved interpretability and reliability compared to skill-agnostic scoring \cite{ye2024flask}. 

Despite these advances, a growing body of meta-evaluation work highlights fundamental limitations of LLM judges. Prior studies document systematic biases such as verbosity and positional bias, limited self-consistency, and sensitivity to prompt design and evaluation protocols \cite{wang2023large, zeng2024llmbar, zheng2023judging, liu2023geval}. In response, recent approaches propose more elaborate judging strategies, including chain-of-thought and decomposition-based evaluation, multi-aspect scoring, reference-based comparisons, and multi-agent or debate-style evaluators such as PRD and ChatEval \cite{gong2023coascore, saha2023branch, li2023prd, chan2024chateval, jeong2024prepair}. However, evidence on the effectiveness of these methods remains mixed. REIFE shows that gains from evaluation protocols depend strongly on the base model and dataset, underscoring the need for diverse and well-calibrated evaluation setups \cite{liu2025reife}. Similarly, Huang et al.\ demonstrate that fine-tuned judge models (e.g., JudgeLM, PandaLM, Auto-J, Prometheus) often fail to generalize beyond their training domain, behaving more like task-specific classifiers than robust evaluators \cite{huang2025empirical}. 

% Taken together, this literature suggests that while LLM-as-judge provides a useful \emph{external} signal of quality, it remains imperfect and potentially unreliable, particularly for complex, multi-step reasoning. This motivates complementary approaches that examine the quality of \emph{intermediate reasoning} and leverage additional signals beyond judge scores—an especially important consideration in multi-agent systems where debate produces rich but open-ended reasoning traces.

\paragraph{Confidence and Uncertainty in LLMs.}
A parallel line of work investigates whether \emph{intrinsic confidence signals} can be used to assess the reliability of LLM outputs. Research in calibration and uncertainty quantification shows that neural probabilities are informative but not inherently well calibrated, meaning that high confidence does not always correspond to correctness \cite{desai2020calibration, kadavath2022language, kuhn2023semantic}. Nevertheless, token-level probabilities remain one of the most direct signals available during generation, and have been widely used to detect hallucinations, factual inconsistencies, and uncertain outputs through log-probability- and entropy-based features \cite{liu2022token, manakul2023selfcheckgpt, mallen2023whennottrust}. 

In addition, work on self-evaluation suggests that LLMs can sometimes produce useful confidence estimates in natural language, though these verbalized signals may diverge from underlying model uncertainty, particularly in multi-step reasoning settings \cite{kadavath2022language, mavi2025selfevaluatingllmsmultisteptasks}. Recent surveys therefore advocate for scalable uncertainty estimation methods that combine intrinsic decoding-time signals with downstream evaluation metrics \cite{kang2025scalable}.

\section{Methodology}
\label{sec:meth}
\begin{figure*}[t]
  \includegraphics[width=\textwidth]{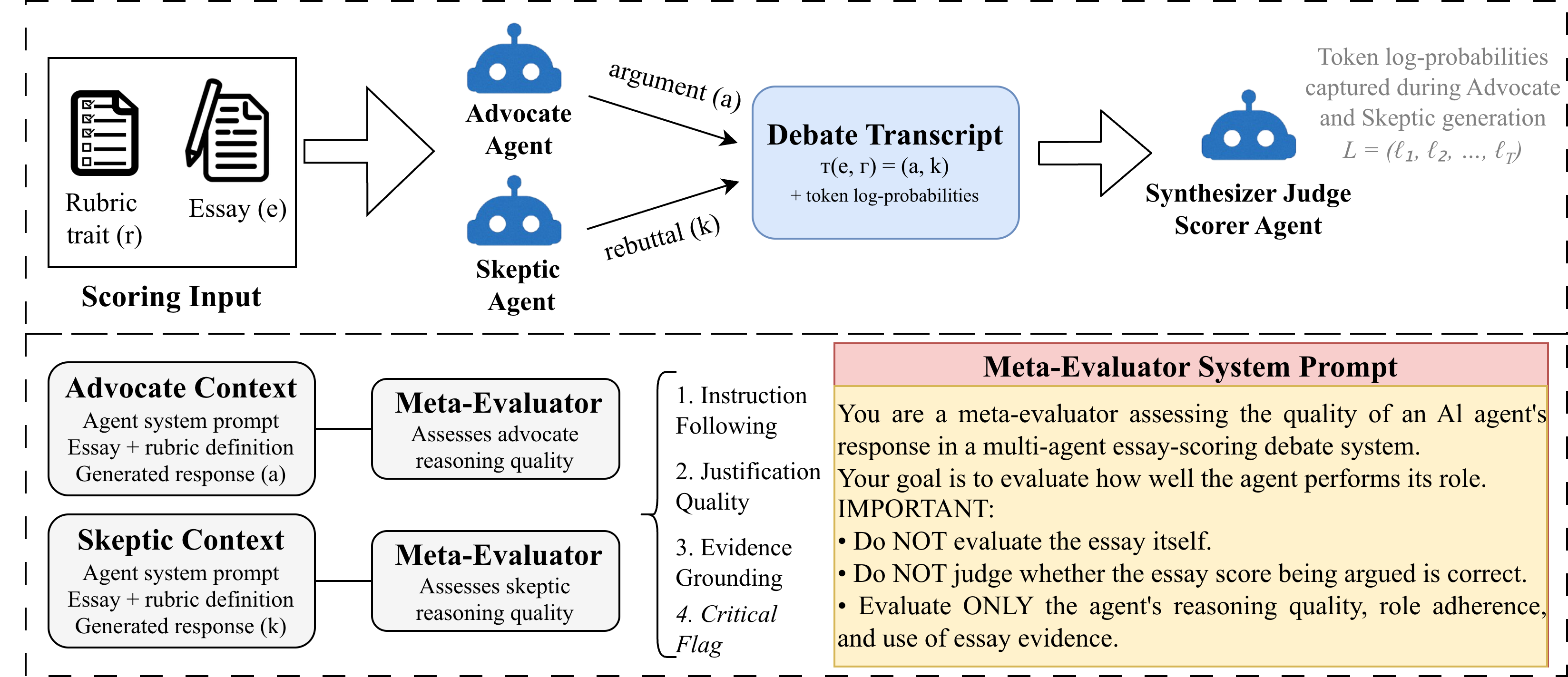}
  \caption{Overview of the proposed multi-agent debate and LLM-as-judge evaluation framework.}
  \label{fig:methodology}
\end{figure*}

Figure~\ref{fig:methodology} provides an overview of our framework, which builds upon the multi-agent debate architecture introduced in prior work~\cite{keramati_2025_17196206} and extends it with an LLM-as-judge meta-evaluation module for reasoning analysis. In the first stage, an \textbf{Advocate} and a \textbf{Skeptic} produce opposing arguments for a given essay–rubric pair while exposing token-level log-probabilities as intrinsic confidence signals. In the second stage, a separate meta-evaluator scores each argument along rubric-based dimensions such as instruction following, justification quality, and evidence grounding. This design enables systematic analysis of the relationship between internal confidence signals and externally judged reasoning quality.

\subsection{Problem Setting}

Let $\mathcal{E}$ denote the set of essays and $\mathcal{R}$ the set of rubric traits. Each essay $e \in \mathcal{E}$ consists of unstructured text together with optional metadata, and each rubric trait $r \in \mathcal{R}$ specifies a textual description and a scoring range $[m_r, M_r]$. For each essay–trait pair $(e, r)$, the debate system produces a transcript
\[
\tau(e,r) = (a,\, k),
\]
where $a$ is the Advocate's argument and $k$ is the Skeptic's rebuttal. Both arguments are generated by a language model conditioned on the essay, rubric, and conversation history; the model simultaneously produces token-level log-probabilities reflecting its internal confidence over candidate continuations.

Given a collection of debate responses $\{(a_i, k_i)\}$, each paired with confidence signals $c_i$ and meta-evaluation scores $q_i$, our objective is to analyze whether token-level probability signals correlate with the externally judged quality of agent reasoning, and thus whether intrinsic confidence can serve as an indicator of reasoning reliability in multi-agent LLM systems.

\subsection{Agents and Roles}

The debate framework comprises three specialized agents that interact sequentially for each essay–trait pair. The \textbf{Advocate} initiates the debate by constructing an argument that highlights the essay's strengths relative to the rubric trait, drawing exclusively on supporting evidence from the essay text without assigning a score. The \textbf{Skeptic} responds by identifying limitations or shortcomings in the essay with respect to the same criterion, producing an evidence-based rebuttal that challenges the Advocate's claims, again without scoring. The \textbf{Synthesizer-Judge Scorer} reads the completed transcript and produces the final trait-level score within the allowed rubric range. Because this agent performs a constrained decision-making task whose output can be evaluated directly against ground-truth scores using accuracy-based metrics, it falls outside the scope of the present study. Our analysis focuses exclusively on the open-ended reasoning produced by the Advocate and Skeptic. Full system prompts for all three agents are provided in \textbf{Appendix~\ref{app:prompts}}.

\subsection{Confidence Signals from Token Log-Probabilities}

We estimate model confidence using token-level log-probabilities obtained during generation. For a generated response of $T$ tokens, the model produces a log-probability at each decoding step:
\[
\ell_t = \log p(t_t \mid t_{<t},\, x),
\]
where $x$ is the prompt context and $t_{<t}$ the preceding tokens. The resulting sequence $L = (\ell_1, \dots, \ell_T)$ forms a log-probability trajectory over the full response.

\subsubsection{Window-Based Segmentation}

Rather than summarizing $L$ with a single statistic, we extract contiguous sub-sequences to examine how confidence evolves across different phases of generation. We use two complementary strategies:

\paragraph{Fixed-length windows.} For window size $k$:
$$
W_{\text{first}}(k) = (\ell_1, \dots, \ell_k), \qquad
$$
$$
W_{\text{last}}(k)  = (\ell_{T-k+1}, \dots, \ell_T).
$$

\paragraph{Percentage-based windows.} To normalize across responses of varying length, we define windows as a fraction $\alpha \in (0,1]$ of the total response:
$$
W_{\text{first}}(\alpha) = (\ell_1, \dots, \ell_{\lfloor \alpha T \rfloor}), 
$$
$$
W_{\text{last}}(\alpha)  = (\ell_{T - \lfloor \alpha T \rfloor + 1}, \dots, \ell_T).
$$

\subsubsection{Statistical Aggregation}

For each window $W$, we compute the following summary statistics:

\paragraph{Mean and median.}
\[
\mu_W = \frac{1}{|W|}\sum_{\ell \in W} \ell, 
\qquad 
\tilde{\mu}_W = \mathrm{median}(W).
\]
The mean reflects overall token likelihood; the median provides a robust central-tendency estimate less sensitive to outlier tokens.

\paragraph{Minimum and maximum.} $\min(W)$ and $\max(W)$ bound the range of token confidence within the segment.

\paragraph{Variance, standard deviation, and range.}
$$
\sigma_W^2 = \frac{1}{|W|}\sum_{\ell \in W}(\ell - \mu_W)^2,
$$
$$
\text{range}_W = \max(W) - \min(W).
$$

These statistics quantify the dispersion and volatility of the generation process within a window.

\paragraph{Trajectory slope.} We fit a linear regression to the log-probability sequence over the window:
\[
\ell_t \approx a\,t + b.
\]
The slope coefficient $a$ captures directional trends: $a > 0$ indicates growing confidence across the segment, while $a < 0$ indicates declining confidence.

\subsection{LLM-as-Judge Meta-Evaluation}
Because the Advocate and Skeptic generate open-ended argumentative reasoning rather than discrete labels, their outputs cannot be evaluated with reference-based metrics such as accuracy or n-gram overlap. We therefore introduce a secondary evaluation stage in which a separate language model judges the quality of each agent's reasoning along rubric-based dimensions.

\subsubsection{Prompt Reconstruction}
For each agent response, we reconstruct the complete prompt context the agent originally received, consisting of: (i) the agent's system instructions describing its role and behavioral constraints, (ii) the rubric trait definition, (iii) the essay text, and (iv) the agent's generated response. Supplying this full context enables the evaluator to assess both role adherence and the appropriateness of the evidence used.

\subsubsection{Evaluation Dimensions}
The meta-evaluator scores each response along three dimensions:
\paragraph{Instruction Following.} Whether the agent maintained its assigned role throughout and avoided prohibited behaviors.
\paragraph{Justification Quality.} Whether claims are supported by explicit reasoning that coherently links evidence to conclusions.
\paragraph{Evidence Grounding.} Whether the argument references concrete, specific passages from the essay rather than relying on vague or generic statements.

\subsubsection{Scoring Protocol}

Each dimension is scored on a three-point ordinal scale ($1 = \text{Low},\ 2 = \text{Medium},\ 3 = \text{High}$), assigned independently. The evaluator also raises a \textbf{critical issue flag} when the response contains a severe failure that invalidates the reasoning, including hallucinated evidence, major internal contradictions, role-constraint violations, or incoherent output.

We summarize reasoning quality using an aggregate score:
\[
Q_1 = s_{\text{instruction}} + s_{\text{justification}} + s_{\text{evidence}}.
\]

If a critical failure is detected, the aggregate score is overridden:
\[
Q =
\begin{cases}
0 & \text{if critical issue is present}, \\
Q_1 & \text{otherwise}.
\end{cases}
\]

This formulation ensures that responses containing severe reasoning failures are penalized regardless of their dimension-level scores, yielding a composite score in the range $[0, 9]$.

\section{Experiments}

\subsection{Experimental Setup}

\paragraph{Data.} We evaluate our framework on the ASAP\footnote{https://www.kaggle.com/c/asap-aes/data} dataset, a widely used benchmark of student-written English essays scored by trained human raters against prompt-specific rubrics. Although ASAP comprises eight essay sets, analytic trait-level annotations are available only for Essay Sets~7 and~8; all experiments are therefore conducted on these two sets, which provide multiple independent human ratings per essay at the trait level. Full dataset statistics, rubric descriptions, and label construction details are provided in Appendix~\ref{app:data}.

\paragraph{Evaluation Metrics.} For ordinal evaluation targets—instruction following, justification quality, evidence grounding, and aggregate score—we report Spearman's $\rho$ and Kendall's $\tau$ to capture rank-order agreement between confidence proxies and LLM-as-judge scores. For the binary \texttt{critical flag}, we report AUROC and point-biserial correlation. To keep results interpretable, we report only the best-performing proxy per target–role combination.

\paragraph{Models.} Advocate and Skeptic responses are generated using \textbf{GPT-4o-mini}, and meta-evaluation is performed by \textbf{GPT-5-mini} instance acting as the LLM-as-judge. To reduce run-to-run variance, the meta-evaluator decodes deterministically, while the Advocate and Skeptic use standard sampling. Token-level log-probabilities are collected during Advocate and Skeptic decoding to compute the confidence proxies described in Section \ref{sec:meth}.

\subsection{Cross-Dataset Analysis}

\begin{table*}[t]
\centering
\small
\begin{tabular}{l l l l c}
\toprule
Set & Role & Target & Top confidence features (ranked) & Best score \\
\midrule
7 & Advocate & Aggregate     & Full-response median, Final-half median, Full-response mean & 0.379 \\
7 & Advocate & Instruction   & Full-response median, Final-half median, Full-response mean & 0.394 \\
7 & Advocate & Justification & Full-response median, Final-half median, Full-response mean & 0.353 \\
7 & Advocate & Evidence      & Full-response median, Final-half median, Full-response mean & 0.242 \\
7 & Advocate & Critical      & Max of first 3 tokens, Max of first 5 tokens, Max of first 10 tokens & 0.849$^\dagger$ \\
\midrule
7 & Skeptic  & Aggregate     & Range of first 3 tokens, Full-response slope, Slope over first 30 tokens & 0.208 \\
7 & Skeptic  & Instruction   & Range of first 3 tokens, Full-response slope, Slope over first 30 tokens & 0.163 \\
7 & Skeptic & Justification & Range of first 3 tokens, Slope over first 30 tokens, Full-response slope & 0.266 \\
7 & Skeptic  & Evidence      & Range of first 3 tokens, Slope over first 30 tokens, Full-response slope & 0.114 \\
7 & Skeptic  & Critical      & Mean of first 3 tokens, Min of first 3 tokens, Slope over first 3 tokens & 0.638$^\dagger$ \\
\midrule
8 & Advocate & Aggregate     & Range of first 3 tokens, Range of first 5 tokens, Std.\ dev.\ of first 3 tokens & 0.320 \\
8 & Advocate & Instruction   & Range of first 3 tokens, Std.\ dev.\ of first 3 tokens, Variance of first 3 tokens & 0.373 \\
8 & Advocate & Justification & Range of first 3 tokens, Range of first 5 tokens, Final-half mean & 0.284 \\
8 & Advocate & Evidence      & Range of first 3 tokens, Range of first 5 tokens, Std.\ dev.\ of first 5 tokens & 0.336 \\
8 & Advocate & Critical      & Median of first 5 tokens, Range of first 3 tokens, Std.\ dev.\ of first 3 tokens & 0.759$^\dagger$ \\
\midrule
8 & Skeptic  & Aggregate     & Range of first 3 tokens, Std.\ dev.\ of first 3 tokens, Variance of first 3 tokens & 0.196 \\
8 & Skeptic  & Instruction   & Range of first 3 tokens, Std.\ dev.\ of first 3 tokens, Variance of first 3 tokens & 0.176 \\
8 & Skeptic  & Justification & Range of first 3 tokens, Std.\ dev.\ of first 3 tokens, Variance of first 3 tokens & 0.231 \\
8 & Skeptic  & Evidence      & Range of first 3 tokens, Std.\ dev.\ of first 3 tokens, Variance of first 3 tokens & 0.092 \\
8 & Skeptic  & Critical      & Median of first 5 tokens, Mean of first 3 tokens, Std.\ dev.\ of first 3 tokens & 0.630$^\dagger$ \\
\bottomrule
\end{tabular}
\caption{Top-3 confidence features for each role--target pair across Essay Sets 7 and 8. For ordinal targets, the final column reports Spearman correlation; for critical detection, it reports AUROC. Early-$k$ refers to statistics computed over the first $k$ generated tokens, while final-half refers to the last 50\% of the response. Full rankings and additional metrics are provided in Appendix~\ref{app:proxy-rankings}. $^\dagger$AUROC reported for critical detection.}
\label{tab:cross-dataset-top3}
\end{table*}
Table~\ref{tab:cross-dataset-top3} summarizes the top-performing confidence features across Essay Sets~7 and~8. While both datasets exhibit a consistent relationship between token-level confidence and judged reasoning quality, the structure of this relationship varies notably across roles and datasets.

For the Advocate, the dominant signal shifts from global to local confidence. In Essay Set~7, the strongest predictors are full-response summaries such as \emph{full-response median} and \emph{final-half median}, which consistently lead across all ordinal targets. In contrast, Essay Set~8 shows a clear transition toward early-generation dispersion, with \emph{range of first 3 tokens} emerging as the top feature across all ordinal targets. This shift suggests that the informativeness of confidence signals depends on dataset characteristics, with some settings favoring globally stable confidence while others emphasize variability at the start of generation.

In contrast, the Skeptic exhibits a stable pattern across both datasets. Early-window dispersion features—particularly \emph{range of first 3 tokens}—consistently dominate all ordinal targets, with only minor variation in secondary features such as slope-based measures. Correlation magnitudes are uniformly lower than for the Advocate, indicating a weaker alignment between confidence and judged quality for adversarial reasoning.

A similar pattern holds for critical failure detection. Advocate failures are best captured by sharp early-token signals, such as \emph{max of first 3 tokens} in Set~7 (AUROC $= 0.849$) and \emph{median of first 5 tokens} in Set~8 (AUROC $= 0.759$), suggesting that severe errors manifest as localized confidence spikes or drops early in generation. Skeptic detection performance is both weaker and more stable across datasets (AUROC $\approx 0.63$), with early mean- and median-based features performing best.

Taken together, three findings are consistent across both essay sets. First, early-generation signals are broadly informative: even when not dominant (as in Advocate Set~7), they remain among the top-performing features across nearly all role--target pairs. Second, the opening phase of generation is the most diagnostically useful region, aligning with trajectory analyses that show higher variability in early tokens compared to later segments. Third, the Advocate--Skeptic asymmetry is robust: confidence aligns more strongly with supportive reasoning than with adversarial critique, both in ordinal correlations and in critical-failure detection. Full rankings and additional metrics are reported in Appendix~\ref{app:proxy-rankings}.

\subsection{Trajectory Analysis}

To understand why early-window features consistently dominate across both datasets, we analyze token-level log-probability trajectories for Advocate and Skeptic responses. Because responses vary in length, each trajectory is normalized to a 0\%–100\% position scale via interpolation. For each role and dataset, we compute the mean trajectory along with 25–75 and 10–90 percentile bands across responses.

Figure~\ref{fig:four-images} reveals a consistent structural pattern across both essay sets and roles: responses begin with relatively high confidence, followed by a sharp early decline, a prolonged mid-response plateau, and a modest recovery toward the end. Since log-probabilities closer to zero indicate higher confidence, this pattern suggests that initial tokens are easy to predict, uncertainty increases as the model transitions into substantive reasoning, and confidence stabilizes once the response structure is established.

A central observation is that variability is concentrated at the beginning of the response. The percentile bands are widest in the first few tokens, indicating substantial heterogeneity in early-generation behavior: some responses start with stable, high-confidence trajectories, while others exhibit immediate volatility. In contrast, the middle and later portions of the response are comparatively flat and tightly clustered. This explains why early-window dispersion features (e.g., \emph{range of first 3 tokens}) consistently emerge as strong predictors—they capture precisely the region where responses differ most in confidence. Once generation reaches the plateau phase, trajectories become too similar for full-response or late-window features to remain discriminative.

The trajectories also reveal a persistent role asymmetry. Skeptic responses exhibit slightly lower average confidence and wider low-confidence tails throughout generation, particularly in the lower percentile bands. This aligns with the weaker correlations observed for Skeptic features and suggests that adversarial reasoning introduces greater variability in generation dynamics. In contrast, Advocate responses follow more stable trajectories, making confidence a more reliable signal of judged quality.

\begin{figure*}[t]
\centering
\includegraphics[width=0.48\linewidth]{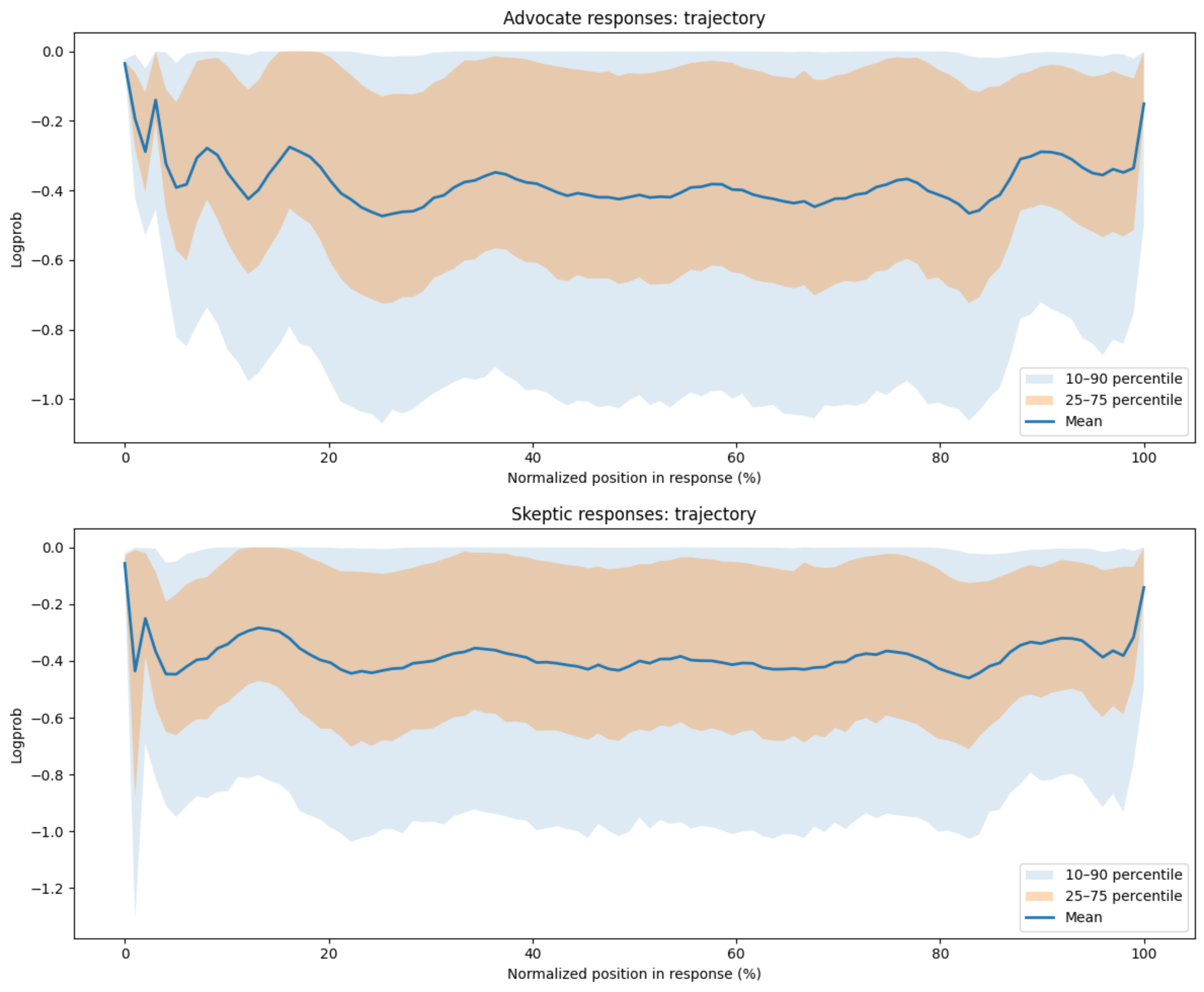}\hfill
\includegraphics[width=0.48\linewidth]{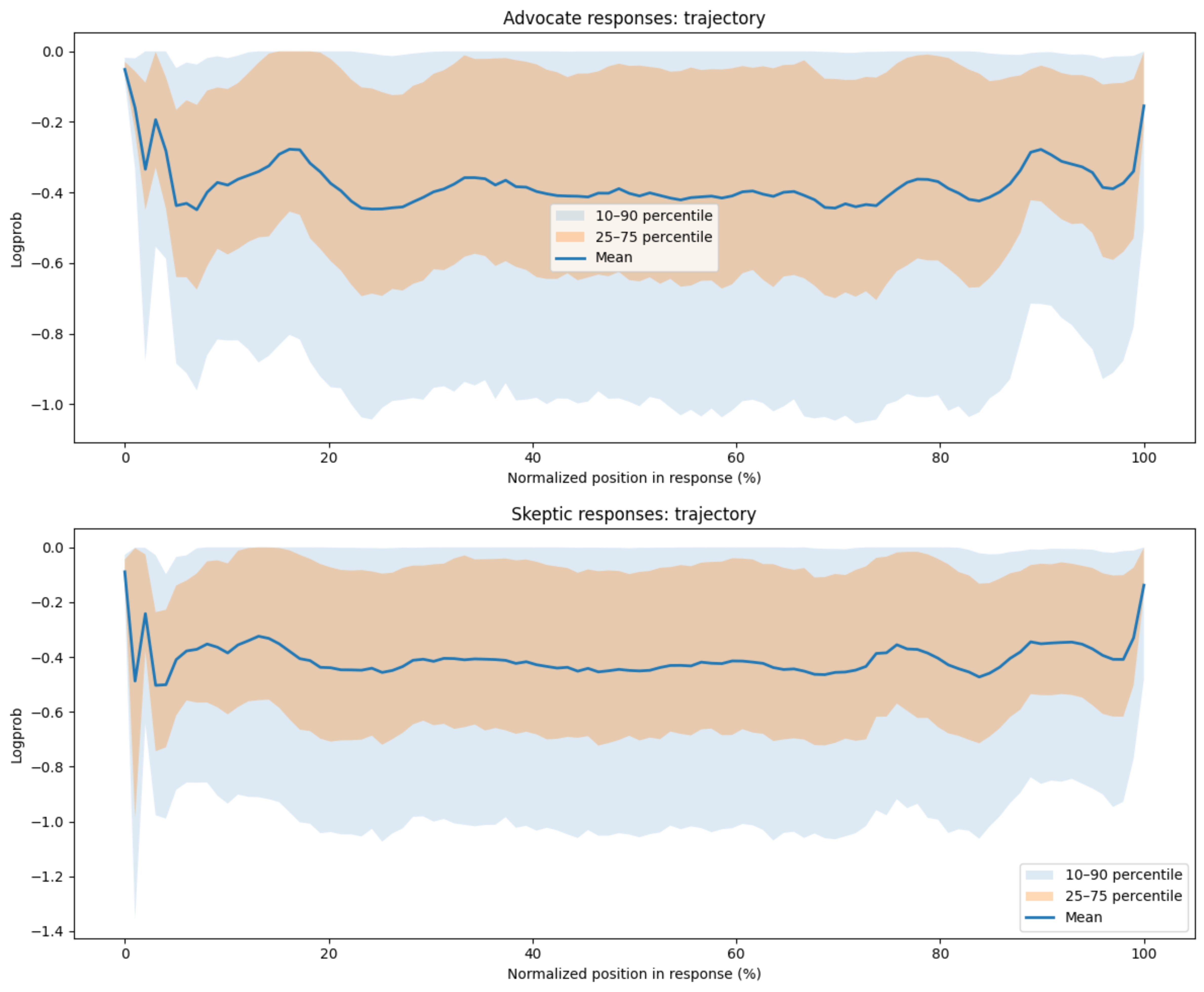}
\caption{Token-level log-probability trajectories for Advocate and Skeptic responses on Essay Sets 7 (left) and 8 (right). Solid lines denote mean trajectories; shaded regions indicate percentile bands (25--75 and 10--90).}
\label{fig:four-images}

% \caption{Normalized token-level log-probability trajectories for Advocate and Skeptic responses in Essay Set 7. Each response is interpolated onto a shared 0--100\% response axis. The solid line shows mean logprob, while shaded bands indicate the 25--75 and 10--90 percentile ranges. Both roles exhibit a high-confidence opening, a sharp early decline, a stable middle region, and a late recovery. The widest spread occurs near the beginning of the response, indicating that early-generation behavior is the most heterogeneous and informative segment.}
\label{fig:four-images}
\end{figure*}

\subsection{Failure Analysis}

\begin{figure*}[t]
  \centering
  \includegraphics[width=\textwidth]{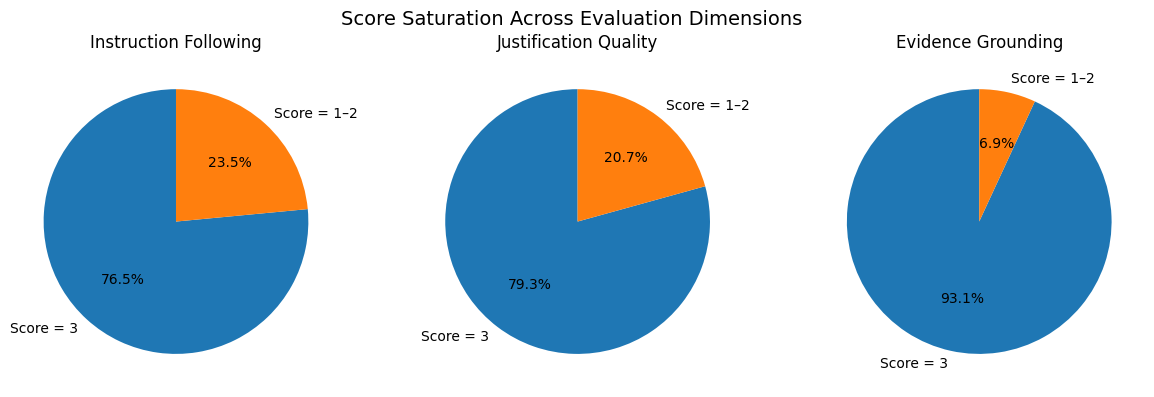}
  \caption{Distribution of LLM-as-judge scores across evaluation dimensions. Evidence grounding is highly saturated at the maximum score, while instruction following and justification quality exhibit greater variability.}
  \label{fig:pie}
\end{figure*}
\paragraph{LLM-as-Judge Meta-Evaluator.}
Figure~\ref{fig:pie} shows that the meta-evaluator exhibits strong score concentration across all dimensions, assigning the maximum score (3) in the majority of cases: 76.5\% for instruction following, 79.3\% for justification quality, and 93.1\% for evidence grounding. Despite this overall skew, the three dimensions differ markedly in their ability to discriminate between responses. Evidence grounding is highly saturated and behaves almost as a binary signal, contributing little variation. In contrast, instruction following and justification quality account for most of the observable differences in scores.

Among these, justification quality shows the strongest role asymmetry, with a 13\% gap in pass rates between the Advocate and Skeptic. Instruction following captures broader procedural failures across both roles, with the Skeptic penalized more heavily (84.2\% vs.\ 93.6\%), primarily due to violations of explicit instructions.

These patterns reflect fundamentally different failure modes across roles. Advocate failures are primarily associated with justification quality, especially through overstatement. The Advocate frequently amplifies weak or incorrect evidence, presenting flawed reasoning as strong or mischaracterizing surface-level features. Importantly, these errors occur along a continuum, ranging from mild exaggeration to clear misrepresentation. 

In contrast, Skeptic failures are dominated by instruction-following violations. The most common issue arises from engaging with anonymization placeholders (e.g., \texttt{@CAPS1}, \texttt{@NUM2}) despite explicit instructions to ignore them. Unlike Advocate errors, these failures are largely binary: the Skeptic either adheres to the procedure or violates it.

This distinction explains the persistent role asymmetry observed in the correlation analysis. Because Advocate errors vary continuously, they induce a broader distribution of scores, enabling stronger alignment with confidence signals (e.g., $\rho = 0.373$ for instruction following). Skeptic errors, by contrast, collapse into near binary outcomes, limiting score variability and compressing rank-based correlations regardless of the underlying signal.

A similar pattern appears in critical failure detection. Advocate failures—often driven by hallucinated or fabricated evidence—produce sharper confidence anomalies, leading to stronger detection performance (e.g., AUROC $\approx 0.76$). These errors typically involve claims about nonexistent essay features, introducing low-probability tokens during generation. In contrast, Skeptic failures are predominantly procedural and do not produce comparable confidence deviations, resulting in weaker detection signals (AUROC $\approx 0.63$).

\section{Conclusion}
We presented a framework that couples multi-agent debate with LLM-as-judge meta-evaluation to study whether token-level confidence signals can predict the quality of open-ended argumentative reasoning in LLM-based essay scoring. Across both ASAP essay sets, we find that early-window log-probability statistics, particularly dispersion measures over the first few generated tokens, are consistently the strongest predictors of externally judged reasoning quality. This finding is supported both by the correlation analysis and by trajectory-level evidence showing that the opening segment of generation is the most heterogeneous, and therefore the most informative, region of the response. We also identify a stable Advocate–Skeptic asymmetry: confidence proxies correlate more reliably with Advocate reasoning quality than with Skeptic reasoning quality, a difference traceable to the distinct failure modes of each role.

\section*{Limitations}
This study is exploratory in scope and several factors limit the generalizability of the findings. Our experiments are conducted on only two ASAP essay sets and a single model family, although these subsets were selected because they are the only portions of the dataset that provide the trait-level annotations required for fine-grained reasoning analysis. Despite this narrower setting, we observe consistent trajectory-level patterns across both datasets, particularly the predictive value of early-token confidence signals.

In addition, the reported correlations are moderate rather than deterministic, suggesting that token-level confidence should be interpreted as a useful auxiliary signal rather than a standalone measure of reasoning quality. Relatedly, our framework relies on LLM-as-judge evaluation, which may partially reflect alignment between models instead of fully objective reasoning assessment. To mitigate this, we use structured rubric-based scoring, deterministic meta-evaluation, multiple complementary metrics, and detailed trajectory and failure analyses. Nevertheless, future work would benefit from broader cross-domain experiments, human validation studies, stronger statistical testing, and evaluation on additional model families and debate settings.

\section*{Ethical Statement}
This work studies confidence estimation and reasoning evaluation in multi-agent LLM systems for automated essay scoring. Because educational assessment systems may inherit biases present in both datasets and language models, the proposed framework should not be viewed as a replacement for human judgment in high-stakes settings. In addition, LLM-as-judge evaluation and token-level confidence estimates are imperfect proxies for reasoning quality and may reflect model-specific biases or calibration issues. Our goal is therefore not to automate educational decision-making, but to better understand the reliability and interpretability of multi-agent reasoning systems. All experiments were conducted on publicly available benchmark data and commercially available language models.

\section*{Acknowledgments}
This paper is based upon work supported by the National Science Foundation under Grant No. 2315294.

\bibliography{custom}

\appendix

\section{Data and Preprocessing}
\label{app:data}

\subsection{Dataset Selection}

We conduct our analysis on the ASAP (Automated Student Assessment Prize) dataset, a standard benchmark for essay scoring. The dataset consists of eight prompt-specific essay sets with varying genres, scoring rubrics, and grade levels. 

Our study focuses exclusively on \textbf{Essay Sets~7 and~8}, as these are the only subsets that provide \emph{trait-level annotations} from multiple human raters. This property is essential for our setup, since we evaluate reasoning quality at the level of individual rubric traits rather than holistic scores. The remaining essay sets are not used, as they only provide single aggregated scores and therefore do not support fine-grained evaluation.

\subsection{Relevant Dataset Characteristics}

Table~\ref{tab:asap_subset} summarizes the key properties of the two essay sets used in our experiments.

\begin{table}[h]
\centering
\small
\caption{Summary of the ASAP subsets used in this work.}
\label{tab:asap_subset}
\begin{tabular}{lccc}
\toprule
\textbf{Set} & \textbf{Grade} & \textbf{\# Essays} & \textbf{\# Traits} \\
\midrule
7 & 7  & 1,569 & 4 \\
8 & 10 & 723   & 6 \\
\bottomrule
\end{tabular}
\end{table}

These two sets differ in both rubric complexity and score ranges, providing a useful testbed for analyzing how confidence signals interact with reasoning quality under different evaluation conditions.

\subsection{Prompt Context}

Each essay is written in response to a fixed prompt. For completeness, we include simplified versions of the prompts used in the selected sets:

\paragraph{Set 7.}
Students are asked to write a story about patience, either from personal experience or imagination.

\paragraph{Set 8.}
Students are asked to write a true story in which laughter plays an important role.

These prompts define the context in which both the debate agents and the evaluator operate.

\subsection{Text Handling}

The ASAP essays are transcriptions of handwritten student responses. We use the text \emph{as provided}, without any normalization or correction. In particular, spelling errors, grammatical inconsistencies, and informal structures are preserved. 

This choice is important because the evaluation criteria (e.g., evidence grounding and justification) depend on the original textual content, and preprocessing could alter signals that are relevant to both the agents and the evaluator.

\section{Top-3 Confidence Feature Rankings by Dataset}
\label{app:proxy-rankings}

Tables~\ref{tab:set7-top3} and \ref{tab:set8-top3-full} report the top three confidence features for each role--target pair on Essay Sets~7 and~8. For ordinal targets, features are ranked by Spearman correlation with the LLM-as-judge score, with Kendall’s $\tau$ reported as a secondary measure. For critical failure detection, features are ranked by AUROC, with point-biserial correlation (PB) included for completeness.

To improve interpretability, we present features using descriptive names rather than implementation-specific identifiers. Early-$k$ refers to statistics computed over the first $k$ generated tokens, while final-half refers to the last 50\% of the response. Full-response features are computed over the entire generated sequence.

\begin{table*}[t]
\centering
\scriptsize
\begin{tabular}{l l l c c c c}
\toprule
Role & Target & Feature & Spearman & Kendall & AUROC & PB \\
\midrule
Advocate & Aggregate     & Full-response median & 0.379 & 0.295 & --    & -- \\
Advocate &               & Final-half median    & 0.363 & 0.281 & --    & -- \\
Advocate &               & Full-response mean   & 0.353 & 0.273 & --    & -- \\
\midrule
Advocate & Evidence      & Full-response median & 0.242 & 0.196 & --    & -- \\
Advocate &               & Final-half median    & 0.237 & 0.193 & --    & -- \\
Advocate &               & Full-response mean & 0.237 & 0.192 & --    & -- \\
\midrule
Advocate & Instruction   & Full-response median & 0.394 & 0.317 & --    & -- \\
Advocate &               & Final-half median    & 0.383 & 0.308 & --    & -- \\
Advocate &               & Full-response mean   & 0.379 & 0.305 & --    & -- \\
\midrule
Advocate & Justification & Full-response median & 0.353 & 0.284 & --    & -- \\
Advocate &               & Final-half median    & 0.332 & 0.267 & --    & -- \\
Advocate &               & Full-response mean   & 0.323 & 0.260 & --    & -- \\
\midrule
Advocate & Critical      & Max of first 3 tokens  & -- & -- & 0.849 & 0.261 \\
Advocate &               & Max of first 5 tokens  & -- & -- & 0.849 & 0.261 \\
Advocate &               & Max of first 10 tokens & -- & -- & 0.830 & 0.253 \\
\midrule
Skeptic  & Aggregate     & Range of first 3 tokens & 0.208 & 0.166 & -- & -- \\
Skeptic  &               & Full-response slope     & 0.125 & 0.100 & -- & -- \\
Skeptic  &               & Slope over first 30 tokens & 0.121 & 0.097 & -- & -- \\
\midrule
Skeptic  & Evidence      & Range of first 3 tokens & 0.114 & 0.093 & -- & -- \\
Skeptic  &               & Slope over first 30 tokens & 0.067 & 0.055 & -- & -- \\
Skeptic  &               & Full-response slope     & 0.060 & 0.049 & -- & -- \\
\midrule
Skeptic  & Instruction   & Range of first 3 tokens & 0.163 & 0.130 & -- & -- \\
Skeptic  &               & Full-response slope     & 0.102 & 0.081 & -- & -- \\
Skeptic  &               & Slope over first 30 tokens & 0.098 & 0.078 & -- & -- \\
\midrule
Skeptic  & Justification      & Range of first 3 tokens & 0.266 & 0.217 & -- & -- \\
Skeptic  &               & Slope over first 30 tokens & 0.231 & 0.201 & -- & -- \\
Skeptic  &               & Full-response slope     & 0.209 & 0.179 & -- & -- \\
\midrule
Skeptic  & Critical      & Mean of first 3 tokens  & -- & -- & 0.638 & 0.194 \\
Skeptic  &               & Min of first 3 tokens   & -- & -- & 0.635 & 0.177 \\
Skeptic  &               & Slope over first 3 tokens & -- & -- & 0.634 & 0.180 \\
\bottomrule
\end{tabular}
\caption{Top-3 confidence features for Essay Set 7.}
\label{tab:set7-top3}
\end{table*}

\begin{table*}[t]
\centering
\scriptsize
\begin{tabular}{l l l c c c c}
\toprule
Role & Target & Feature & Spearman & Kendall & AUROC & PB \\
\midrule
Advocate & Aggregate     & Range of first 3 tokens & 0.320 & 0.248 & -- & -- \\
Advocate &               & Range of first 5 tokens & 0.274 & 0.214 & -- & -- \\
Advocate &               & Std.\ dev.\ of first 3 tokens & 0.267 & 0.205 & -- & -- \\
\midrule
Advocate & Evidence      & Range of first 3 tokens & 0.336 & 0.272 & -- & -- \\
Advocate &               & Range of first 5 tokens & 0.326 & 0.263 & -- & -- \\
Advocate &               & Std.\ dev.\ of first 5 tokens & 0.305 & 0.247 & -- & -- \\
\midrule
Advocate & Instruction   & Range of first 3 tokens & 0.373 & 0.302 & -- & -- \\
Advocate &               & Std.\ dev.\ of first 3 tokens & 0.338 & 0.274 & -- & -- \\
Advocate &               & Variance of first 3 tokens & 0.338 & 0.274 & -- & -- \\
\midrule
Advocate & Justification & Range of first 3 tokens & 0.284 & 0.228 & -- & -- \\
Advocate &               & Range of first 5 tokens & 0.284 & 0.228 & -- & -- \\
Advocate &               & Final-half mean & 0.273 & 0.221 & -- & -- \\
\midrule
Advocate & Critical      & Median of first 5 tokens & -- & -- & 0.759 & 0.285 \\
Advocate &               & Range of first 3 tokens  & -- & -- & 0.754 & -0.287 \\
Advocate &               & Std.\ dev.\ of first 3 tokens & -- & -- & 0.736 & -0.286 \\
\midrule
Skeptic  & Aggregate     & Range of first 3 tokens & 0.196 & 0.157 & -- & -- \\
Skeptic  &               & Std.\ dev.\ of first 3 tokens & 0.195 & 0.157 & -- & -- \\
Skeptic  &               & Variance of first 3 tokens & 0.195 & 0.157 & -- & -- \\
\midrule
Skeptic  & Evidence      & Range of first 3 tokens & 0.092 & 0.075 & -- & -- \\
Skeptic  &               & Std.\ dev.\ of first 3 tokens & 0.090 & 0.074 & -- & -- \\
Skeptic  &               & Variance of first 3 tokens & 0.090 & 0.074 & -- & -- \\
\midrule
Skeptic  & Instruction   & Range of first 3 tokens & 0.176 & 0.140 & -- & -- \\
Skeptic  &               & Std.\ dev.\ of first 3 tokens & 0.174 & 0.139 & -- & -- \\
Skeptic  &               & Variance of first 3 tokens & 0.174 & 0.139 & -- & -- \\
\midrule
Skeptic  & Justification & Range of first 3 tokens & 0.231 & 0.189 & -- & -- \\
Skeptic  &               & Std.\ dev.\ of first 3 tokens & 0.230 & 0.188 & -- & -- \\
Skeptic  &               & Variance of first 3 tokens & 0.230 & 0.188 & -- & -- \\
\midrule
Skeptic  & Critical      & Median of first 5 tokens & -- & -- & 0.630 & 0.151 \\
Skeptic  &               & Mean of first 3 tokens   & -- & -- & 0.630 & 0.156 \\
Skeptic  &               & Std.\ dev.\ of first 3 tokens & -- & -- & 0.617 & -0.138 \\
\bottomrule
\end{tabular}
\caption{Top-3 confidence features for Essay Set 8.}
\label{tab:set8-top3-full}
\end{table*}

\section{Agent Prompt Templates}
\label{app:prompts}

This section provides the system instructions used to define the roles of the agents in the debate framework. 
All prompts are implemented as template files and rendered at runtime using a shared context dictionary. 
The context includes the rubric trait name, the full rubric definition serialized as JSON, the essay text, 
the essay prompt or question, and the valid scoring range for the trait.

In addition, the Synthesizer-Judge receives the debate transcript produced by the Advocate and Skeptic agents. 
These prompts establish strict role boundaries to ensure that each agent performs a specialized function in the debate process.

The prompt configurations were developed through iterative experimentation, including pilot runs and refinements designed to enforce role adherence, maintain output consistency, and reduce undesired behaviors such as assigning scores prematurely or mixing multiple rubric traits in a single argument.

For transparency and reproducibility, we provide the exact system instructions used to define each agent role.

\subsection{Advocate Agent}

The Advocate agent is responsible for presenting arguments that highlight the strengths of the essay with respect to a single rubric trait. The agent receives the essay text and the rubric definition and produces an evidence-based argument explaining how the essay satisfies the expectations of the trait.

The Advocate is explicitly instructed to focus only on positive aspects of the essay and to avoid assigning a score or discussing weaknesses. The agent may reference specific passages from the essay to support its claims.

\begin{figure*}[h]
\centering
\begin{tcolorbox}[title=\textbf{Advocate Agent System Prompt}, colback=blue!3, colframe=blue!60]

You are an Advocate Agent participating in a multi-agent debate system for essay evaluation.

Your task is to analyze the essay and highlight strengths that demonstrate how the essay satisfies the rubric expectations for the trait "\$TRAIT\_NAME".

Focus exclusively on positive evidence from the essay. Provide detailed reasoning supported by specific excerpts or paraphrased examples from the essay.

Do not assign a score and do not critique weaknesses. Your role is solely to present arguments supporting the essay’s strengths with respect to the specified rubric trait.

Anonymization  
\$ANON\_CONTEXT

\end{tcolorbox}
\caption{System instructions for the Advocate agent.}
\label{fig:prompt_advocate}
\end{figure*}

\subsection{Skeptic Agent}

The Skeptic agent provides a counterargument to the Advocate by identifying weaknesses or limitations in the essay relative to the same rubric trait.

The Skeptic receives both the essay text and the Advocate's argument and produces a critique that challenges the strengths presented or highlights aspects where the essay fails to meet the rubric expectations.

The agent is instructed not to assign a score and to focus exclusively on critical analysis.

\begin{figure*}[h]
\centering
\begin{tcolorbox}[title=\textbf{Skeptic Agent System Prompt}, colback=red!3, colframe=red!60]

You are a Skeptic Agent participating in a multi-agent debate system for essay evaluation.

Your task is to critically analyze the essay and identify weaknesses related to the rubric trait "\$TRAIT\_NAME".

Provide detailed critiques supported by specific references to the essay text. Focus only on identifying shortcomings or areas where the essay does not fully satisfy the rubric expectations.

Do not assign a score and do not discuss strengths. Your role is to challenge the essay’s performance with respect to the specified rubric trait.

Anonymization  
\$ANON\_CONTEXT

\end{tcolorbox}
\caption{System instructions for the Skeptic agent.}
\label{fig:prompt_skeptic}
\end{figure*}

\subsection{Synthesizer-Judge Agent}

The Synthesizer-Judge serves as the final decision-maker in the debate process. This agent reads the arguments produced by the Advocate and Skeptic and determines a final score for the rubric trait.

The agent synthesizes the competing arguments and evaluates them against the rubric definition before assigning a score within the permitted range.

\begin{figure*}[h]
\centering
\begin{tcolorbox}[title=\textbf{Synthesizer-Judge Agent System Prompt}, colback=gray!5, colframe=black!60]

You are the Synthesizer-Judge in a multi-agent debate system for essay evaluation.

Your task is to read the debate transcript between the Advocate and Skeptic agents regarding the rubric trait "\$TRAIT\_NAME".

Carefully consider the arguments presented by both agents and evaluate them against the rubric expectations.

Based on the combined evidence, assign a final integer score between \$MIN\_POINTS and \$MAX\_POINTS for the essay on this rubric trait.

Anonymization  
\$ANON\_CONTEXT

\end{tcolorbox}
\caption{System instructions for the Synthesizer-Judge agent.}
\label{fig:prompt_judge}
\end{figure*}

\subsection{LLM-as-Judge Meta-Evaluator}
\label{sec:meta-evaluator-agent}

The meta-evaluator agent is responsible for assessing the quality of each Advocate and Skeptic response. The agent receives the original system prompt given to agents, context including the essay text and rubric, and the agent's response.

The meta-evaluator scores each response along three dimensions: instruction following, justification quality, and evidence grounding, each on a three-point ordinal scale (1 = Low, 2 = Medium, 3 = High). The evaluator also flags a critical issue when the response contains hallucinated evidence, severe deviation or violation of instructions, or internal contradictions.

The meta-evaluator is explicitly instructed to evaluate only the agent's reasoning quality and role adherence. It does not assess the essay itself or judge whether the score being argued is correct.

The complete system instructions and output schema are provided in Figures~\ref{fig:prompt_advocate}-~\ref{fig:judge_eval_prompt}.

\begin{figure*}[h]
\centering
\begin{tcolorbox}[title=\textbf{Meta-Evaluation System Prompt}, colback=blue!8, colframe=blue!40!black]

You are a meta-evaluator assessing the quality of an AI agent's response in a multi-agent essay-scoring debate system. Your goal is to evaluate how well the agent performs its role.

\textbf{Important:} Do NOT evaluate the essay itself. Do NOT judge whether the essay score being argued is correct. Evaluate ONLY the agent's reasoning quality, role adherence, and use of essay evidence.

You will receive: (1) the agent's system prompt, (2) the task prompt given to the agent, and (3) the agent's response. Evaluate the response across three dimensions using the full range of the scale: \textbf{1 = Low}, \textbf{2 = Medium}, \textbf{3 = High}. Avoid defaulting to the middle score. Evaluate each dimension independently.

\medskip
\textbf{Dimension 1 -- Instruction Following.}
3: Fully maintains role; completes all task components; no deviations.
2: Generally follows instructions with a minor omission or slight deviation.
1: Major or multiple deviations; neglects important instructions.

\medskip
\textbf{Dimension 2 -- Justification Quality.}
3: Multiple claims with clear reasoning; claim $\rightarrow$ explanation $\rightarrow$ implication structure.
2: At least one supported claim; reasoning understandable but shallow or repetitive.
1: Minimal or vague reasoning; assertions without explanation.

\medskip
\textbf{Dimension 3 -- Evidence Grounding.}
3: Two or more precise references including quotes or detailed paraphrases.
2: One clear identifiable reference; other claims rely on general statements.
1: Evidence vague, indirect, or missing.

\medskip
\textbf{Critical Issues Flag.} Set \texttt{critical\_flag = 1} if any of the following occur: hallucinated essay evidence, severe internal contradiction, explicit instruction violation, or nonsensical output. Otherwise \texttt{critical\_flag = 0}.

\medskip
\textbf{Output:} Return only a JSON object with fields \texttt{instruction\_following}, \texttt{justification\_quality}, \texttt{evidence\_grounding}, \texttt{critical\_flag}, \texttt{critical\_issues\_description}, and \texttt{reasoning} (2--3 sentences).

\end{tcolorbox}
\caption{System instructions for the Meta-Evaluator agent.}
\label{fig:judge_system_prompt}
\end{figure*}

\begin{figure*}[h]
\centering
\begin{tcolorbox}[title=\textbf{Meta-Evaluation Task Prompt}, colback=green!7, colframe=green!40!black, coltitle=white, colbacktitle=green!40!black]

\texttt{\# Agent Being Evaluated: \{AGENT\_TYPE\}}

\medskip
\texttt{\# Agent's System Prompt (Instructions Given to the Agent)}\\
\texttt{\{AGENT\_SYSTEM\_PROMPT\}}

\medskip
\texttt{\# Agent's User Prompt (Task Context Given to the Agent)}\\
\texttt{\{AGENT\_USER\_PROMPT\}}

\medskip
\texttt{\# Agent's Actual Response}\\
\texttt{\{AGENT\_RESPONSE\}}

\medskip
\noindent Now evaluate this agent's response. Use the 3-point scale (1=Low, 2=Medium, 3=High) for each scored dimension and set the \texttt{critical\_flag} to 0 or 1. Output ONLY the JSON object.

\end{tcolorbox}
\caption{Task prompt provided to the Meta-Evaluator agent.}
\label{fig:judge_eval_prompt}
\end{figure*}

\end{document}